\documentclass{ijclclp}
\usepackage{hyperref}
\usepackage{xcolor}
\usepackage{caption}
\usepackage{subcaption}
\usepackage{enumitem}
\usepackage{longtable}
\usepackage{float}

\title{The Coin Flip Judge? Reliability and Bias in LLM-as-a-Judge Evaluation}
\author{Abel Yagubyan}
\affiliation{Independent Researcher}
\email{\texttt{abelyagubyan@berkeley.edu}}

\begin{document}
\maketitle
\thispagestyle{firstpage}

\begin{abstract}
LLM-as-a-Judge is now widely used to rank model outputs, train reward models, and populate public leaderboards, but its run-to-run reliability remains under-characterized. We study repeated identical evaluations on 29 tasks spanning 10 categories using two OpenAI judge models (GPT-4o-mini and GPT-4.1-mini), with 50 pairwise trials and 50 pointwise trials per question, supplemented by temperature and prompt-sensitivity ablations. Across judges, pairwise preferences flip on average 13.6\% of the time, with 28\% of questions exceeding a 20\% flip rate and one question reaching 56\%. GPT-4o-mini also exhibits a significant first-position bias (72\% A-majority, $p = 0.024$). At the same time, mean pointwise score gaps are small (0.19--0.36 on a 10-point scale) and not statistically significant in aggregate, producing a pairwise--pointwise gap: judges frequently choose a winner even when their own scalar scores provide little evidence of a meaningful quality difference. Beyond within-judge instability, cross-judge agreement is only 76\% ($\kappa = 0.51$), semantically equivalent prompt templates change majority outcomes in 25\% of tested cases, and deterministic decoding reduces but does not eliminate inconsistency. A reliability-curve analysis shows that, in our dataset, 11 repeated trials are needed for a majority vote to recover the 50-trial reference verdict with 95\% probability on average, rising to 15 for high-variance questions. These findings suggest that single-trial LLM judging is often too noisy for high-stakes evaluation, and that multi-trial aggregation, position randomization, and explicit uncertainty reporting should be standard practice. Because both judges are from a single provider, cross-provider replication remains an important next step.

\vspace{0.5em}
\end{abstract}

\section{Introduction}

Large language model (LLM) judges are now central to modern evaluation pipelines. They are used to rank model outputs, approximate human preferences in benchmark construction, and serve as reward-model proxies in RLHF systems \citep{zheng2023judging,dubois2024alpacafarm,ouyang2022training,bai2022constitutional}. This shift has made automated evaluation dramatically cheaper and more scalable, but it also raises a basic measurement question: \emph{if we ask the same judge the same question multiple times, do we get the same answer?}

That question is distinct from the better-studied issue of whether LLM judges are biased. Prior work has documented position bias, verbosity bias, and self-enhancement effects in single-trial settings \citep{wang2023large,zheng2023judging,stureborg2024large}. Our focus is complementary. We study \emph{repeated-trial reliability}: the extent to which a fixed judge, given identical candidate responses under nominally identical conditions, returns the same verdict across runs. This matters directly for benchmark validity. If pairwise verdicts fluctuate across repeated trials, then single-trial leaderboards and paper tables can be unstable even when the underlying model outputs are held fixed.

We investigate this question on 29 tasks spanning 10 categories using two OpenAI judge models, with 50 pairwise trials and 50 pointwise trials per question, plus temperature and prompt-sensitivity ablations. The paper has three central messages. First, pairwise judging is often noisier than it appears: mean flip rate is 13.6\%, and 28\% of questions exceed 20\% flip rate. Second, pairwise verdicts and pointwise scores can come apart: judges often select a winner even when their own average scalar scores show little evidence of a meaningful quality gap. Third, reliability improves predictably but nonlinearly with repeated voting: in our dataset, 11 trials are needed on average for a majority vote to recover the 50-trial reference verdict with 95\% probability.

More broadly, we argue that LLM-judge reliability is not a single property. It decomposes into at least four layers: stochastic instability within a judge, systematic bias such as first-position preference, protocol sensitivity to prompt wording and temperature, and disagreement across judge models. Treating these as separate layers clarifies why apparently reasonable evaluation pipelines can still produce brittle conclusions.

Our contributions are therefore: (i) a formal framework for separating pairwise verdicts, pointwise scores, intra-judge consistency, and cross-judge agreement; (ii) a repeated-trial empirical study of LLM judge reliability across 29 tasks and 10 categories; (iii) an analysis of the pairwise--pointwise gap, showing that forced-choice verdicts can overstate evidence for quality differences; (iv) a reliability-curve analysis that translates flip-rate estimates into concrete trial-count recommendations; and (v) practical guidance for reporting uncertainty in LLM-as-a-Judge evaluation.

\section{Related Work}

\subsection{LLM-as-a-Judge Frameworks}

\citet{zheng2023judging} introduced MT-Bench and the LLM-as-a-Judge paradigm, demonstrating that GPT-4 judgments correlate well with human preferences in aggregate while identifying position and verbosity biases. \citet{dubois2024alpacafarm} proposed AlpacaFarm for instruction-following evaluation, showing that LLM judges can approximate human annotators at lower cost, though length bias remained a concern. \citet{liu2023geval} proposed G-Eval, using chain-of-thought prompting and probability calibration to improve judge alignment with human judgments; they showed CoT substantially helps but did not study run-to-run variance. \citet{zhu2023judgelm} proposed JudgeLM, a fine-tuned judge model optimized for consistency, and reported improved agreement scores over prompted GPT-4, although single-trial evaluation was used throughout.

\subsection{Systematic Bias in LLM Evaluation}

\citet{wang2023large} conducted a comprehensive study of LLM judge biases, cataloguing position bias, verbosity bias, and self-enhancement tendencies, and proposed a calibration approach (swap-augmented evaluation) to mitigate position effects. Their work tests a \textit{single trial per bias condition}, not repeated sampling; our study is the first to quantify these biases at the 50-trial scale. \citet{stureborg2024large} showed that LLM judges are systematically influenced by superficial features including response length, formatting, and bullet-point density, findings consistent with verbosity bias, but did not measure within-judge run-to-run variance. \citet{shankar2024validates} raised the meta-question of who validates the validators, arguing that LLM-judge frameworks require empirical calibration against human annotators on each target task, a position strongly supported by our findings.

\subsection{Judge Reliability and Calibration}

The reliability of automated evaluation metrics has been studied in the pre-LLM era: \citet{amidei2019agreement} surveyed human inter-annotator agreement for NLG evaluation tasks, reporting $\kappa = 0.3$--$0.6$ for subjective tasks, a range that our cross-judge $\kappa = 0.51$ matches. \citet{clark2021all} showed that human evaluators of generated text exhibit substantial disagreement ($\sim$20\% pairwise inconsistency), situating LLM judge inconsistency in a broader human-evaluation context.

Within the LLM-judge literature, calibration work has focused primarily on reducing \textit{systematic} bias rather than stochastic variance. Our work is orthogonal: we study \textit{random} variance (what changes if you re-run the same evaluation), which is irreducible by bias correction but addressable through multi-trial aggregation. Concurrent and complementary work by \citet{shankar2024validates} and others highlights that the identity of the judge model substantially affects outcomes, consistent with our $\kappa = 0.51$ inter-judge finding.

\subsection{Reliability Measurement in Psychometrics}

Intraclass correlation (ICC) is the standard psychometric reliability coefficient for repeated-measures designs \citep{amidei2019agreement}. ICC values below 0.60 are conventionally classified as ``poor to moderate'' reliability \citep{clark2021all}, providing a principled interpretation framework for our ICC(2,1) estimates. The use of majority voting to aggregate stochastic classifiers is well-studied in ensemble learning; our reliability curve analysis (Section~\ref{sec:reliability}) provides the first such analysis for LLM judge aggregation, showing that the gain from additional trials follows a concave curve with diminishing returns beyond $\sim$20 trials.

\subsection{Positioning This Work}

Our study is most directly comparable to \citet{stureborg2024large}, who also measure LLM judge inconsistency. Key differences: (i) we use 50 trials per question (vs.\ $\leq$5 in most prior work), enabling high-precision flip rate estimates with bootstrap confidence intervals; (ii) we introduce the pairwise-pointwise paradox as a distinct failure mode, showing that pairwise forced-choice amplifies non-existent quality differences; (iii) we provide a reliability curve and ICC analysis grounded in psychometric methodology; (iv) we quantify the downstream impact via a leaderboard noise budget. Our work has complementary scope to \citet{wang2023large} (who study systematic bias with response swapping) and extends it with stochastic variance analysis.

\section{Formal Framework}

We distinguish four related but non-identical layers of LLM-as-a-Judge behavior: (i) the \emph{pairwise verdict} produced when a judge is forced to choose between two responses, (ii) the \emph{pointwise score} assigned when each response is evaluated independently, (iii) the \emph{intra-judge consistency} of repeated evaluations by the same judge under fixed conditions, and (iv) the \emph{cross-judge agreement} between different judge models on the same items. This separation is important because instability in any one layer can undermine benchmark validity even when the others appear well-behaved.

\textbf{Definition 1 (Judge Evaluation Trial).} A judge evaluation trial is a stochastic mapping from an input tuple $(q, r_A, r_B, p, \theta)$ to an output, where $q$ is the prompt or question, $r_A$ and $r_B$ are candidate responses, $p$ is an evaluation prompt template, and $\theta$ denotes judge-side settings such as model choice, decoding temperature, and response order. In pairwise mode the output is $y \in \{A, B, \mathrm{tie}\}$; in pointwise mode the output is a scalar score $s \in [1,10]$.

\textbf{Definition 2 (Intra-Judge Consistency).} For a fixed tuple $(q, r_A, r_B, p, \theta)$ and judge model $J$, intra-judge consistency is the stability of the output distribution across repeated trials. Perfect consistency means all repeated trials yield the same verdict (pairwise) or the same score (pointwise); lower consistency corresponds to a wider repeated-trial distribution.

\textbf{Definition 3 (Flip Rate).} For $N$ repeated pairwise trials with outcome counts $(n_A, n_B, n_{\mathrm{tie}})$, the flip rate is
\begin{equation}
    \mathrm{FR} = 1 - \frac{\max(n_A, n_B, n_{\mathrm{tie}})}{N}.
\end{equation}
This measures the fraction of trials not supporting the majority outcome. Higher flip rate indicates greater pairwise instability.

\textbf{Definition 4 (Pairwise--Pointwise Gap).} Let $\bar{s}_A$ and $\bar{s}_B$ denote mean pointwise scores across repeated trials. The pairwise--pointwise gap refers to the empirical situation in which pairwise verdicts appear decisive while the corresponding pointwise score gap $|\bar{s}_A - \bar{s}_B|$ is small or statistically indistinguishable from zero.

This framework motivates three hypotheses tested in the paper:
\begin{enumerate}[nosep]
    \item \textbf{H1: Pairwise instability exceeds what pointwise score gaps alone would predict.} In particular, many questions with small mean score gaps will still exhibit non-trivial pairwise flip rates.
    \item \textbf{H2: Position bias varies systematically across judges.} Even under randomized presentation, some judges will exhibit stronger first-position preference than others.
    \item \textbf{H3: Consensus reliability follows a concave saturation curve.} Additional trials improve majority-vote reliability quickly at first, then with diminishing returns.
\end{enumerate}

\section{Methodology}

\subsection{Evaluation Dataset}

We construct a diverse evaluation set of 29 question-response pairs spanning 10 categories: writing (3), reasoning (3), coding (3), knowledge (3), math (3), roleplay (2), extraction (3), ethics (2), instruction-following (3), and hard/ambiguous tasks (4). For each question, we use two high-quality responses from different model tiers (GPT-4o-mini and GPT-4o) to ensure meaningful comparison targets.

Response pairs are deliberately chosen to be competitive, with both responses high-quality but differing in style, structure, or approach. Pointwise evaluation confirms this: across both judges, Response A averages $9.3/10$ ($\sigma = 0.9$) and Response B averages $9.4/10$ ($\sigma = 0.6$), indicating that both responses are consistently rated as high-quality. This design maximizes the sensitivity of our consistency measurements; trivially different responses would yield artificially high consistency. We note that this represents a stress test: real-world evaluation often involves more diverse quality levels, and consistency may be higher for response pairs with obvious quality differences.

\subsection{Judge Models}

We evaluate two judge models from the GPT-4 family:
\begin{itemize}[nosep]
    \item \textbf{GPT-4o-mini}: A cost-efficient model commonly used for large-scale evaluation
    \item \textbf{GPT-4.1-mini}: A newer variant from the GPT-4.1 family
\end{itemize}

Both models are accessed through the OpenAI API. The main experiment uses the default temperature ($t=1.0$) to reflect real-world usage; a supplementary ablation study evaluates $t=0$ (deterministic decoding).

\subsection{Evaluation Protocol}

\textbf{Experiment 1 (Main).} For each (judge, question) pair, we conduct:

\begin{enumerate}[nosep]
    \item \textbf{Pairwise comparison} ($\times 50$): The judge is asked ``Which response is better?'' with randomized A/B presentation order across trials.
    \item \textbf{Pointwise scoring} ($\times 50$ per response): Each response is independently scored on a 1--10 scale.
\end{enumerate}

This yields $29 \times 2 \times (50 + 50 + 50) = 8{,}700$ total API calls. The 50-trial design provides sufficient statistical power to distinguish genuine preferences from noise (binomial test $p < 0.05$ requires $\geq 33/50$ for significance).

\textbf{Experiment 2 (Temperature Ablation).} We repeat the pairwise comparison with $t=0$ for 10 trials per (judge, question) pair, yielding an additional $29 \times 2 \times 10 = 580$ API calls. While $t=0$ should theoretically produce deterministic outputs, API-level factors (batching, quantization, floating-point nondeterminism) may introduce variation.

\textbf{Experiment 3 (Prompt Sensitivity).} We design a second, semantically equivalent prompt template with different framing and structure, and evaluate 10 diverse questions (one per category) with 20 trials per (judge, prompt, question) combination, yielding $10 \times 2 \times 2 \times 20 = 800$ additional API calls.

\subsection{Metrics}

\textbf{Flip Rate.} For each question, we define the flip rate as:
\begin{equation}
    \text{FR} = 1 - \frac{\max(n_A, n_B, n_{\text{tie}})}{N}
\end{equation}
where $n_A$, $n_B$, $n_{\text{tie}}$ are the counts of each outcome across $N$ trials. A flip rate of 0\% indicates perfect consistency; 50\% indicates random behavior (i.e., the minority vote approaches the majority vote). This measures outcome \textit{uncertainty}: a question with FR $= 0.14$ is one where 14\% of trials would yield the non-majority verdict.

\textbf{Outcome Entropy.} As a complementary measure, we report the Shannon entropy of the outcome distribution:
\begin{equation}
    H = -\sum_{o \in \{A, B, \text{tie}\}} p_o \log_2 p_o
\end{equation}
Entropy $H = 0$ bits indicates a deterministic outcome; $H = \log_2 3 \approx 1.58$ bits indicates a uniform distribution over all three outcomes. Entropy captures outcome spread more fully than flip rate, which ignores tie frequency.

\textbf{Position Bias Index.} We measure the fraction of questions where response A (presented first) wins the majority vote:
\begin{equation}
    \text{PBI} = \frac{|\{q : \text{majority}(q) = A\}|}{|\mathcal{Q}|}
\end{equation}
An unbiased judge would yield PBI $\approx 0.5$. We test significance using a sign test.

\textbf{Pairwise-Pointwise Gap.} For each question, we compute:
\begin{equation}
    \text{PPG} = |\bar{s}_A - \bar{s}_B|
\end{equation}
where $\bar{s}_A$ and $\bar{s}_B$ are mean pointwise scores across 50 trials. We test whether aggregate pointwise scores differ using the Wilcoxon signed-rank test.

\textbf{Intraclass Correlation.} We compute ICC(2,1) (two-way random effects, absolute agreement, single measures) from the 50 repeated pointwise scores per response per judge. ICC(2,1) treats each trial as a ``rater'' and each question-response pair as a ``subject,'' measuring the proportion of total variance attributable to genuine quality differences between subjects versus stochastic noise. ICC $< 0.60$ is conventionally classified as poor to moderate reliability.

\textbf{Cross-Judge Agreement.} We report both raw agreement percentage and Cohen's $\kappa$ to account for chance agreement.

\textbf{Reliability Curve.} We simulate using only $K$ randomly sampled trials (Monte Carlo, 500 repetitions per $K$) and compute $P(\text{majority}(K\text{ trials}) = \text{majority}(50\text{ trials}))$ for $K = 1, \ldots, 50$. This characterizes how quickly the majority verdict stabilizes, and provides a principled basis for trial count recommendations.

All confidence intervals are computed via bootstrap resampling (10,000 iterations).

\subsection{Statistical Analysis}

We report nonparametric tests whenever the relevant distributions are small-sample, skewed, or clearly non-Gaussian. Category comparisons use Kruskal--Wallis tests; aggregate pointwise comparisons use the Wilcoxon signed-rank test; position-bias significance is assessed with a sign test on question-level majorities; and cross-judge agreement is summarized with both raw agreement and Cohen's $\kappa$. Because this is an exploratory reliability study with several related outcome measures, we emphasize effect magnitudes and confidence intervals alongside $p$-values rather than treating binary significance as the only decision criterion. ICC(2,1) is used for repeated pointwise scores because it measures absolute agreement under a two-way random-effects design, matching our interpretation of repeated judge calls as interchangeable raters. A linear mixed-effects formulation would be a natural extension for larger future datasets, but is unnecessary for the current descriptive analysis.

\section{Results}

\subsection{Intra-Judge Consistency}

We begin with the most direct repeated-trial question: how often does the same judge change its pairwise verdict when nothing about the evaluated responses changes? At the aggregate level, both judges exhibit mean flip rates near 14\%, but this average conceals substantial heterogeneity across questions.

\begin{table}[ht]
\centering
\caption{Summary of judge consistency metrics across 29 evaluation tasks ($t=1.0$). FR = flip rate, PBI = position bias index, PPG = pairwise--pointwise gap. 95\% confidence intervals are bootstrap intervals.}
\label{tab:summary}
\begin{tabular}{lcc}
\toprule
\textbf{Metric} & \textbf{GPT-4o-mini} & \textbf{GPT-4.1-mini} \\
\midrule
Mean FR (\%) & 13.3 [7.7, 19.4] & 13.9 [7.7, 20.3] \\
Median FR (\%) & 4.0 & 6.0 \\
Max FR (\%) & 46.0 & 56.0 \\
Questions with FR $>$ 20\% & 8/29 (28\%) & 8/29 (28\%) \\
Questions with FR $=$ 0\% & 8/29 (28\%) & 10/29 (34\%) \\
PBI (A majority) & 72\% ($p=0.024$) & 59\% ($p=0.458$) \\
Mean PPG & 0.19 & 0.36 \\
\bottomrule
\end{tabular}
\end{table}

\begin{figure}[H]
    \centering
    \includegraphics[width=\linewidth]{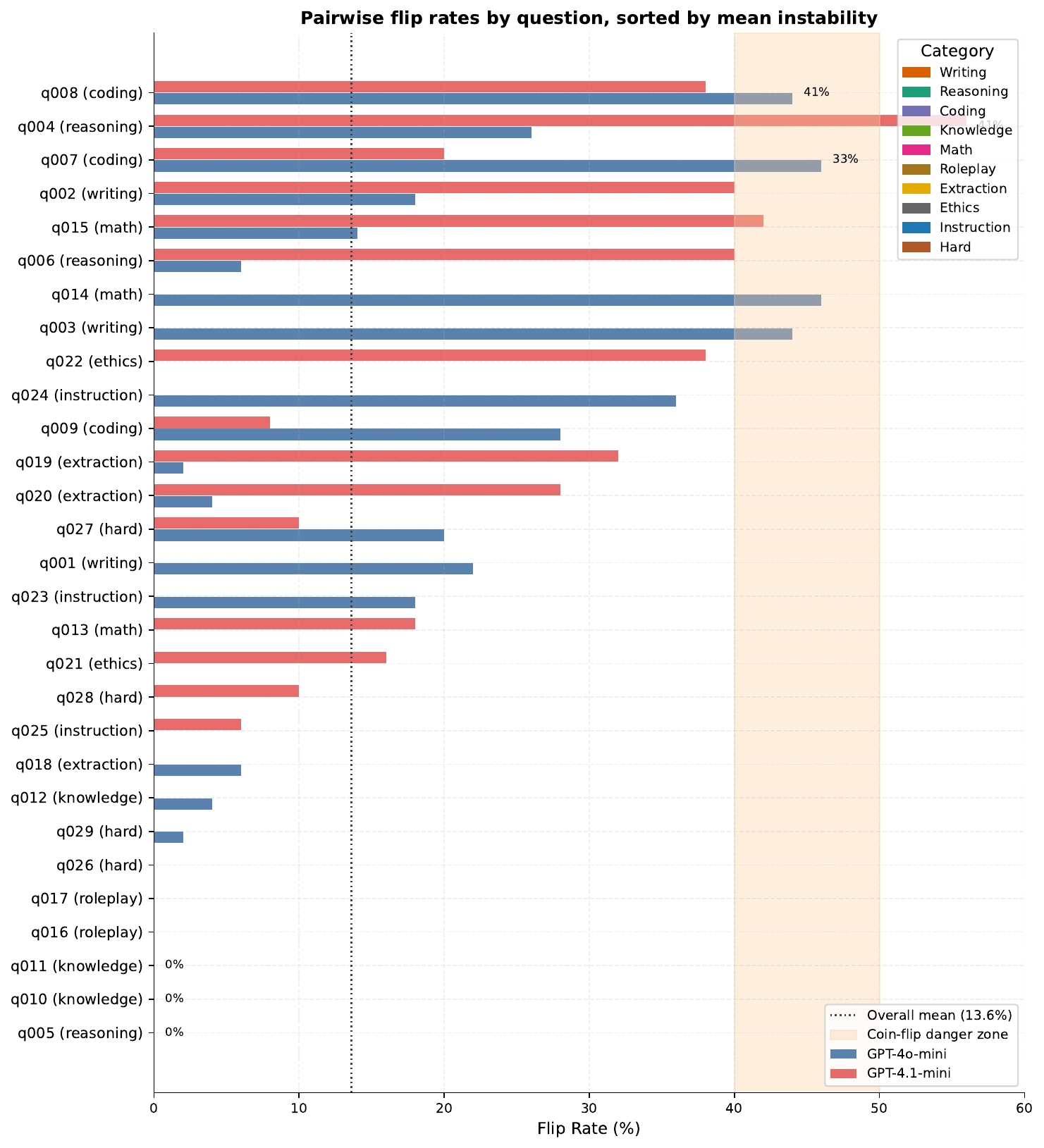}
    \caption{Pairwise preference flip rates across all 29 questions, sorted by mean instability across judges. Each question appears as a paired horizontal bar (one per judge), making it easier to compare where the two judges agree on stability and where they diverge. The shaded band marks the 40--50\% ``coin-flip danger zone,'' where repeated evaluation becomes highly unstable.}
    \label{fig:flip_rates}
\end{figure}

Table~\ref{tab:summary} presents aggregate consistency statistics. Both judges exhibit mean flip rates of approximately 14\%, indicating that roughly one in seven evaluations would change if re-run. Relative to the random baseline of 50\%, the observed mean flip rate is much lower (Cohen's $d = 3.07$ in magnitude), so the judges are clearly not random overall. But the distribution is highly bimodal (Figure~\ref{fig:flip_rates}): many questions show near-perfect consistency (FR $= 0$\%), while others approach coin-flip territory. The two judges do not differ meaningfully in overall flip rate (Mann-Whitney $U = 438$, $p = 0.783$; $\eta^2 \approx 0.0003$ for judge identity on item-level flip rates).

The most notable finding is the existence of extreme inconsistency: GPT-4.1-mini reaches a 56\% flip rate on q004 (a reasoning task), meaning that on this item the majority verdict is unstable enough to be worse than a fair 50/50 split. Eight questions per judge exceed 20\% flip rates, concentrated in coding, writing, and reasoning. This already supports H1: instability is not a marginal phenomenon confined to a few pathological cases, but a recurring feature of competitively matched evaluation items.

\subsection{Position Bias}

\begin{table}[ht]
\centering
\caption{Position-bias summary by judge. ``A'' denotes the first-presented response.}
\label{tab:position_bias}
\begin{tabular}{lccc}
\toprule
\textbf{Judge} & \textbf{A majorities} & \textbf{B/tie majorities} & \textbf{Sign-test $p$} \\
\midrule
GPT-4o-mini & 21/29 (72\%) & 8/29 (28\%) & 0.024 \\
GPT-4.1-mini & 17/29 (59\%) & 12/29 (41\%) & 0.458 \\
\bottomrule
\end{tabular}
\end{table}

Table~\ref{tab:position_bias} shows substantial position bias. GPT-4o-mini displays a strong primacy effect, with the first-presented response (A) winning majority preference in 21 of 29 questions (72\%, sign test $p = 0.024$). GPT-4.1-mini is more balanced at 59\% ($p = 0.458$, not significant). This between-judge difference is substantively important even though the study includes only two judges, and supports H2: position bias is not a fixed property of the evaluation protocol alone, but also of the judge model. At the individual question level, 24 of 29 questions show significant position bias ($p < 0.05$, binomial test) for \textit{both} judges, indicating that position effects are pervasive even when aggregate bias appears moderate.

This position bias has direct implications for evaluation fairness: a model whose response happens to appear first may systematically receive more favorable evaluations.

\subsection{The Pairwise-Pointwise Paradox}
\label{sec:results_paradox}

\begin{figure}[H]
    \centering
    \includegraphics[width=0.85\linewidth]{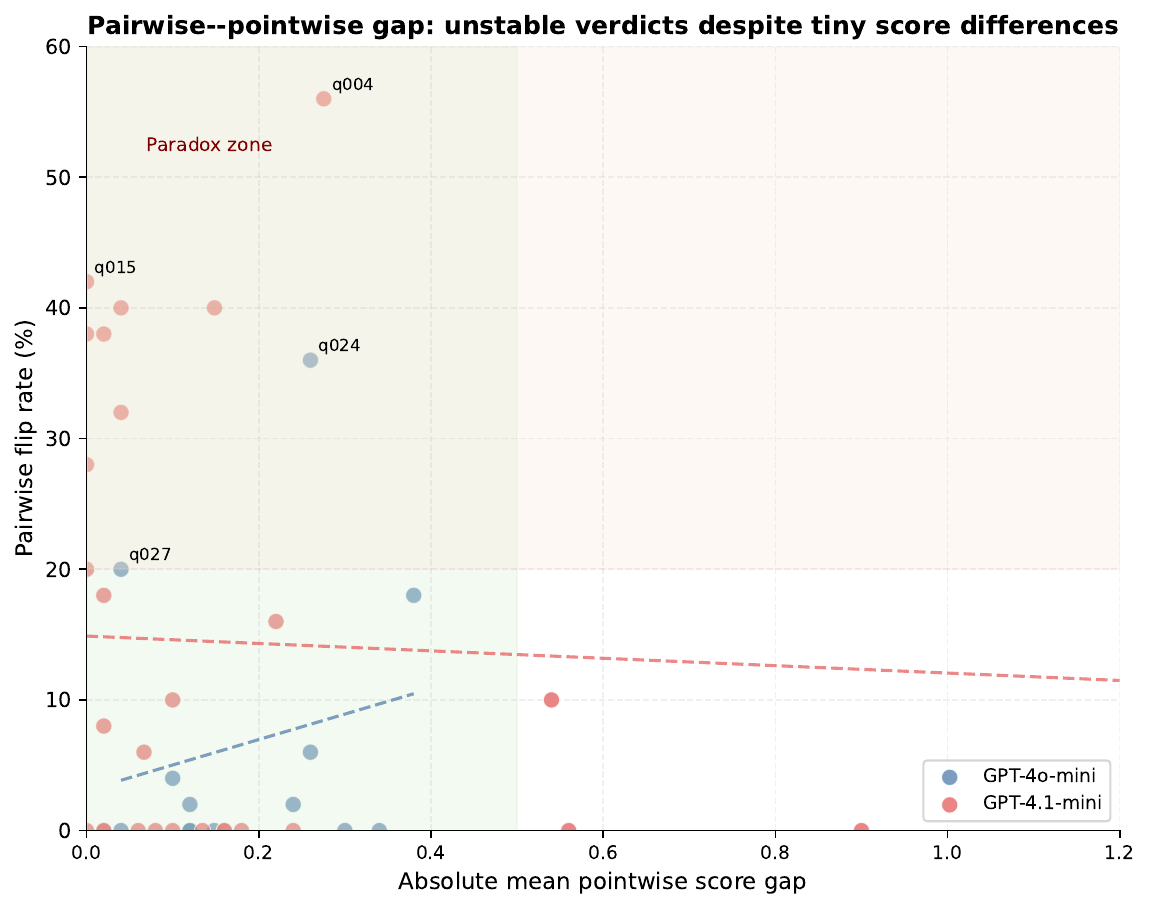}
    \caption{Pairwise--pointwise gap. Many questions with very small mean pointwise score gaps still exhibit substantial pairwise flip rates. The shaded lower-left region highlights the paradox zone: little scalar evidence of a quality difference, yet unstable forced-choice verdicts.}
    \label{fig:paradox}
\end{figure}

Perhaps the most notable finding is the disconnect between pairwise preferences and pointwise scores. The mean pointwise score gap is only 0.19 points for GPT-4o-mini and 0.36 for GPT-4.1-mini on a 10-point scale. Aggregate pointwise scores do \textit{not} significantly differ between responses A and B for either judge (Wilcoxon signed-rank: GPT-4o-mini $W=42$, $p=0.827$; GPT-4.1-mini $W=96.5$, $p=0.126$). Yet in pairwise mode, the same judges still pick ``winners.''

Figure~\ref{fig:paradox} illustrates this pairwise--pointwise gap. Questions with near-zero score differences often exhibit high flip rates, suggesting that forced-choice evaluation can amplify weak or nonexistent scalar preferences into unstable ordinal verdicts. This supports H1 directly: repeated pairwise behavior is often more decisive in form than in evidential content. In practical terms, a pairwise winner should not automatically be interpreted as evidence of a robust underlying quality difference.

\subsection{Category Analysis}

\begin{figure}[H]
    \centering
    \includegraphics[width=\linewidth]{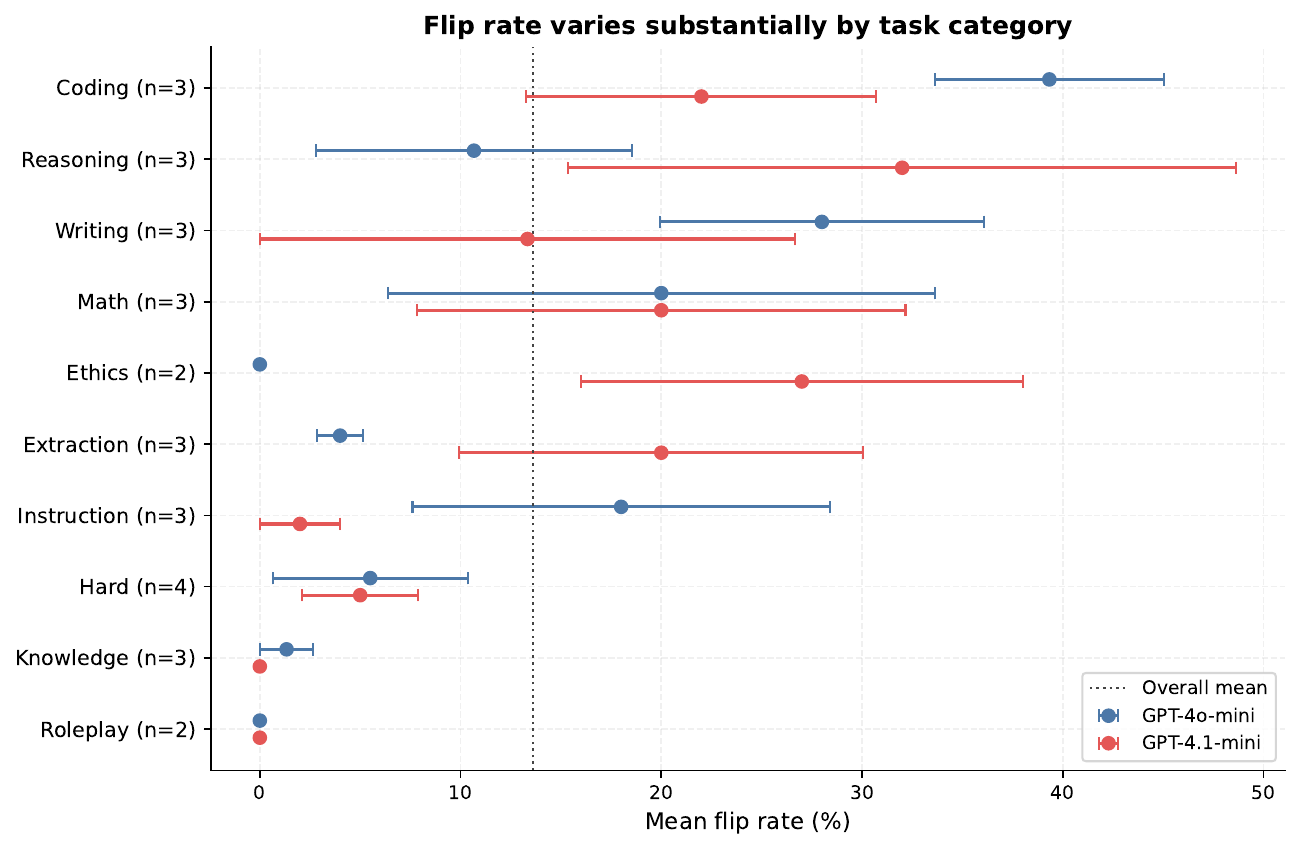}
    \caption{Mean flip rate by task category, shown as point estimates with standard-error bars and ordered from highest to lowest mean instability. Category effects are visually large even though judge-specific significance tests are underpowered.}
    \label{fig:categories}
\end{figure}

Figure~\ref{fig:categories} shows flip rates broken down by task category. There is a visible trend toward higher inconsistency in subjective categories (coding, writing, reasoning) compared to factual ones (knowledge, roleplay), although the judge-specific Kruskal--Wallis tests do not reach conventional significance thresholds (GPT-4o-mini $H=15.8$, $p=0.071$; GPT-4.1-mini $H=11.4$, $p=0.252$), likely due to the small number of questions per category. Still, category explains a non-trivial share of item-level flip-rate variance overall ($\eta^2 \approx 0.31$), indicating that task type matters much more than judge identity in this dataset.

The pattern differs substantially between judges:

\begin{itemize}[nosep]
    \item \textbf{Coding}: High inconsistency for GPT-4o-mini (39\%) but moderate for GPT-4.1-mini (22\%)
    \item \textbf{Reasoning}: Moderate for GPT-4o-mini (11\%) but high for GPT-4.1-mini (32\%)
    \item \textbf{Knowledge/Roleplay}: Consistently low flip rates ($<$5\%) for both judges
    \item \textbf{Ethics}: Stable for GPT-4o-mini (0\%) but variable for GPT-4.1-mini (27\%)
\end{itemize}

This category-dependent inconsistency suggests that judge reliability varies substantially by task type, and that the \textit{pattern} of unreliability differs across judges, a practically important point for evaluation pipelines that rely on a single judge.

\subsection{Difficulty-Stratified Analysis}

To assess whether inconsistency is primarily driven by ambiguous questions, we stratify questions by difficulty, defined as the mean flip rate across both judges. Questions with mean FR $< 10\%$ are classified as ``easy'' (clear winner), while those with FR $\geq 10\%$ are ``hard'' (ambiguous).

\begin{table}[h]
\centering
\caption{Flip rates stratified by question difficulty. Easy questions show near-deterministic behavior; hard questions exhibit substantial instability.}
\label{tab:difficulty}
\begin{tabular}{lccc}
\toprule
\textbf{Stratum} & \textbf{N} & \textbf{Mean FR} & \textbf{Categories} \\
\midrule
Easy (FR $<$ 10\%) & 14 & 2.9\% & knowledge, roleplay, extraction \\
Hard (FR $\geq$ 10\%) & 15 & 23.6\% & coding, writing, reasoning \\
\bottomrule
\end{tabular}
\end{table}

Table~\ref{tab:difficulty} reveals a strongly bimodal distribution: nearly half of questions (14/29) are judged with high consistency (mean FR = 2.9\%), while the remaining 15 questions exhibit substantial instability (mean FR = 23.6\%). The easy--hard contrast is large (Cohen's $d = 2.96$), indicating that instability is concentrated rather than diffuse. Easy questions cluster in factual and well-defined categories (knowledge, roleplay, extraction), while hard questions concentrate in subjective or open-ended categories (coding, writing, reasoning). This suggests that LLM judge inconsistency is not uniformly distributed but rather concentrated in task types where evaluation criteria are inherently more subjective, a pattern also observed in human annotation studies \citep{amidei2019agreement}.

\subsection{Cross-Judge Agreement}

Within-judge stability is only one part of evaluation reliability. Even if each judge were internally stable, benchmark conclusions could still vary if different judge models systematically disagree. We therefore analyze cross-judge agreement separately from intra-judge inconsistency.

\begin{figure}[H]
    \centering
    \includegraphics[width=\linewidth]{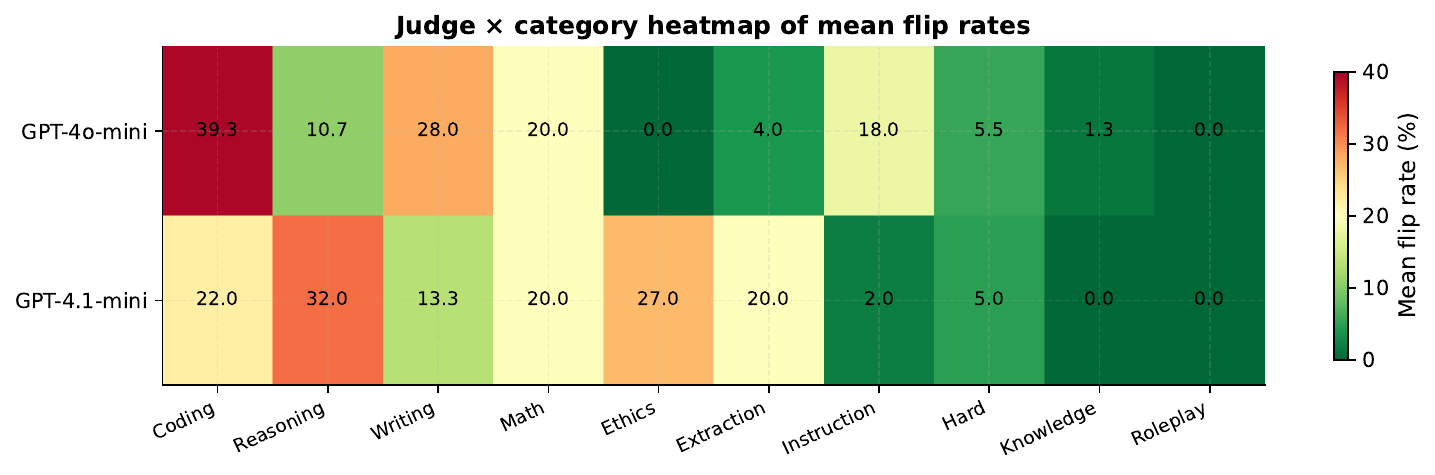}
    \caption{Judge $\times$ category heatmap of mean flip rates. Compared with per-question bar charts, this view emphasizes a cleaner structural point: task type explains substantially more variation in instability than judge identity does in this dataset.}
    \label{fig:heatmap}
\end{figure}

The two judges agree on the majority-preferred response for only 22 of 29 questions (76\%), yielding Cohen's $\kappa = 0.51$ (moderate agreement). Disagreements are concentrated in writing (q002, q003), coding (q007, q008, q009), and hard tasks (q028). In three cases, GPT-4.1-mini declares a tie while GPT-4o-mini picks a winner, suggesting different decision thresholds.

For context, this $\kappa = 0.51$ is comparable to the lower end of human inter-annotator agreement on subjective NLG tasks ($\kappa = 0.3$--$0.6$; \citealt{amidei2019agreement}) and notably below the 81\% agreement reported for MT-Bench human evaluators \citep{zheng2023judging}. The 76\% inter-judge agreement rate means that approximately one in four evaluation outcomes depends on which judge model is selected.

\subsection{Temperature Ablation}

One obvious mitigation for stochastic inconsistency is to reduce decoding randomness. We therefore test whether setting temperature to zero removes repeated-trial variance, or merely attenuates it.

\begin{figure}[H]
    \centering
    \includegraphics[width=\linewidth]{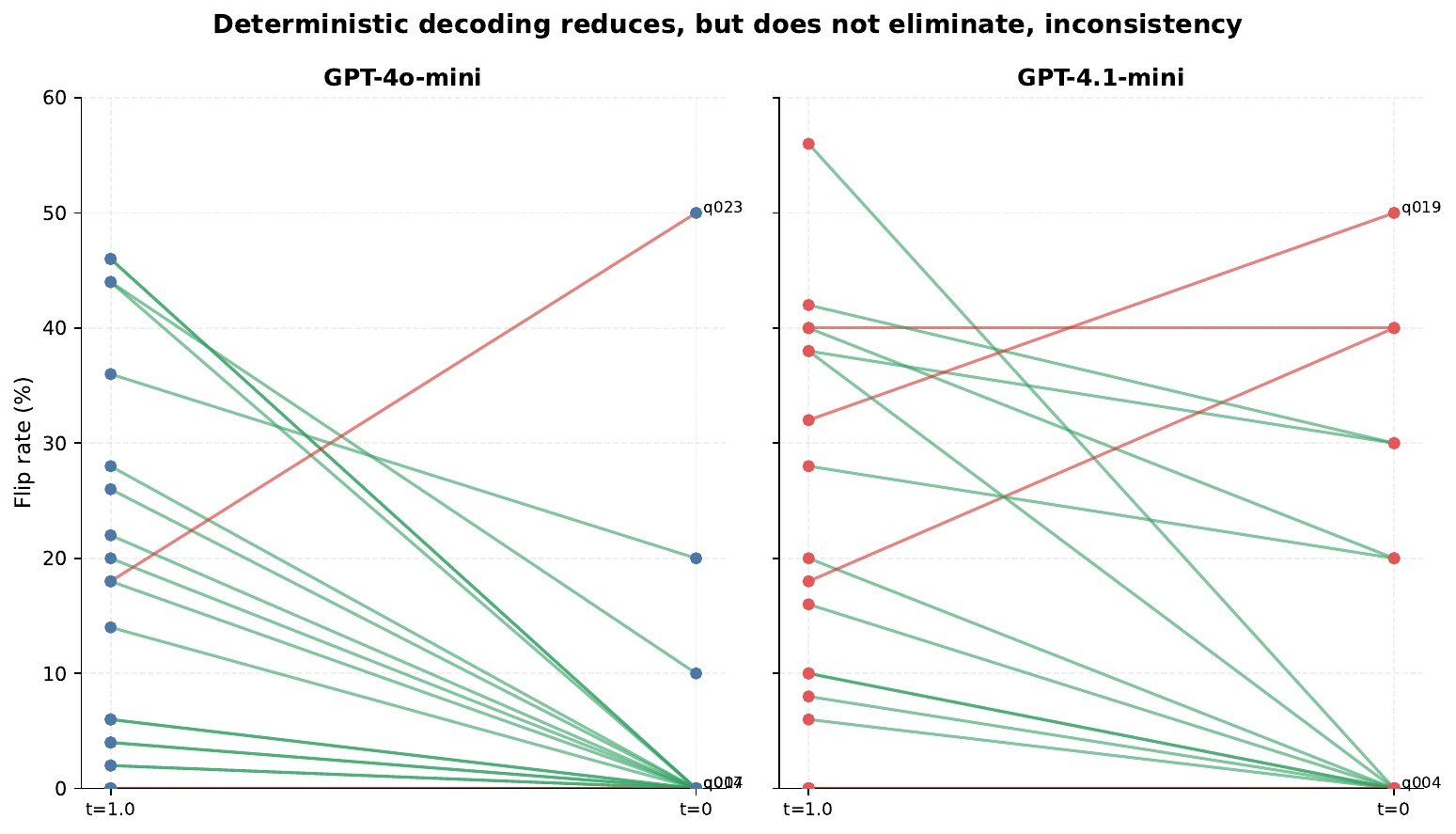}
    \caption{Temperature ablation as a slopegraph from $t=1.0$ to $t=0$. Most questions move downward, showing that deterministic decoding reduces instability, but several retain non-zero flip rates even at $t=0$, especially for GPT-4.1-mini.}
    \label{fig:temp}
\end{figure}

\begin{table}[h]
\centering
\caption{Temperature ablation results. Flip rates at $t=0$ vs $t=1.0$.}
\label{tab:temp}
\begin{tabular}{lccc}
\toprule
\textbf{Judge} & \textbf{FR ($t=1.0$)} & \textbf{FR ($t=0$)} & \textbf{Reduction} \\
\midrule
GPT-4o-mini & 13.3\% & 2.8\% & 79\% \\
GPT-4.1-mini & 13.9\% & 7.9\% & 43\% \\
\bottomrule
\end{tabular}
\end{table}

Figure~\ref{fig:temp} and Table~\ref{tab:temp} present the temperature ablation results. Setting $t=0$ substantially reduces flip rates for GPT-4o-mini (79\% reduction, from 13.3\% to 2.8\%) but is less effective for GPT-4.1-mini (43\% reduction, to 7.9\%). Even at $t=0$, GPT-4.1-mini exhibits non-zero flip rates on 7 of 29 questions, with one reaching 50\%.

This residual inconsistency at $t=0$ likely reflects API-level nondeterminism (floating-point variation, batch-processing effects) rather than intentional sampling. It demonstrates that \textbf{deterministic decoding is necessary but not sufficient} for consistent evaluation, and that additional strategies (multi-trial voting, multi-judge panels) remain useful even when temperature is controlled.

\subsection{Prompt Template Sensitivity}

Prompt wording is often treated as a minor implementation detail in LLM-as-a-Judge studies. Here we treat it as part of the evaluation protocol itself. If semantically equivalent prompt templates yield meaningfully different verdicts, then ``the judge'' is not just the model, but the model-prompt pair.

\begin{figure}[H]
    \centering
    \includegraphics[width=\linewidth]{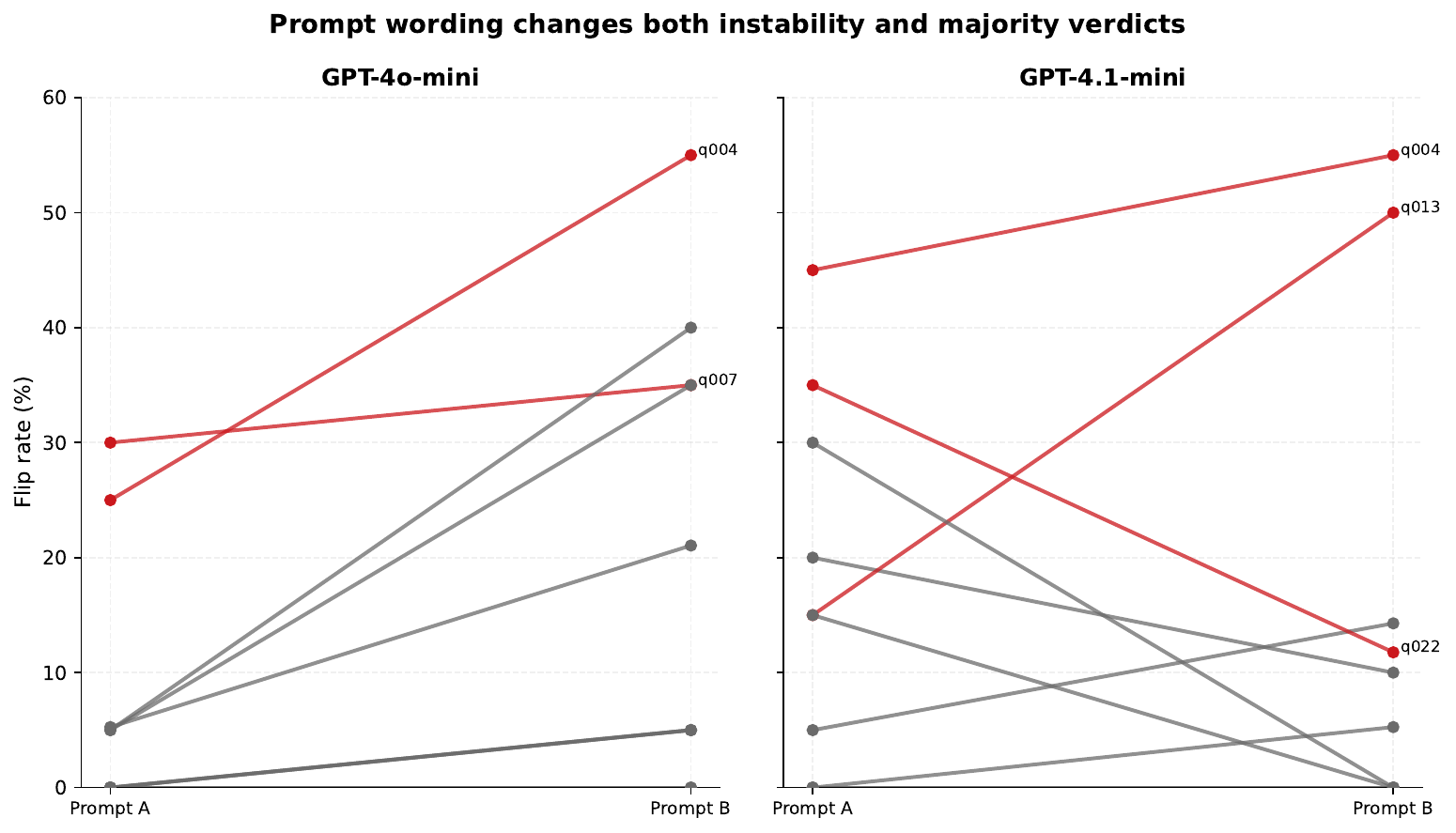}
    \caption{Prompt sensitivity shown as prompt-to-prompt slopegraphs on a 10-question subset. Changes in wording affect both flip rate and, for several items, the majority verdict itself, reinforcing that prompt template is a real experimental variable rather than a cosmetic choice.}
    \label{fig:prompt}
\end{figure}

To assess sensitivity to prompt wording, we designed two semantically equivalent but stylistically different evaluation prompts and tested them on a 10-question subset (20 trials each, both judges). Prompt A uses our standard format (``You are an impartial judge...''), while Prompt B uses an alternative framing (``Please act as a fair and unbiased evaluator...'' with step-by-step instructions).

\begin{table}[h]
\centering
\caption{Prompt template sensitivity on a 10-question subset. Cross-prompt agreement measures whether the majority-preferred response is the same under both prompts.}
\label{tab:prompt}
\begin{tabular}{lccc}
\toprule
\textbf{Judge} & \textbf{Cross-prompt} & \textbf{Mean $\Delta$FR} \\
 & \textbf{agreement} & \\
\midrule
GPT-4o-mini & 8/10 (80\%) & 11.6\% \\
GPT-4.1-mini & 7/10 (70\%) & 15.2\% \\
Combined & 15/20 (75\%) & 13.4\% \\
\bottomrule
\end{tabular}
\end{table}

As shown in Figure~\ref{fig:prompt} and Table~\ref{tab:prompt}, changing the prompt template flips the majority-preferred response in 25\% of cases (5/20), with an average absolute change in flip rate of 13.4 percentage points. Prompt B generally induces higher inconsistency, particularly for questions that were already borderline under Prompt A. This finding shows that evaluation outcomes are sensitive not only to temperature and judge model, but also to the specific wording of the evaluation prompt, an often-overlooked source of variance.

\subsection{Reliability as a Function of Trial Count}
\label{sec:reliability}

\begin{figure}[H]
    \centering
    \includegraphics[width=\linewidth]{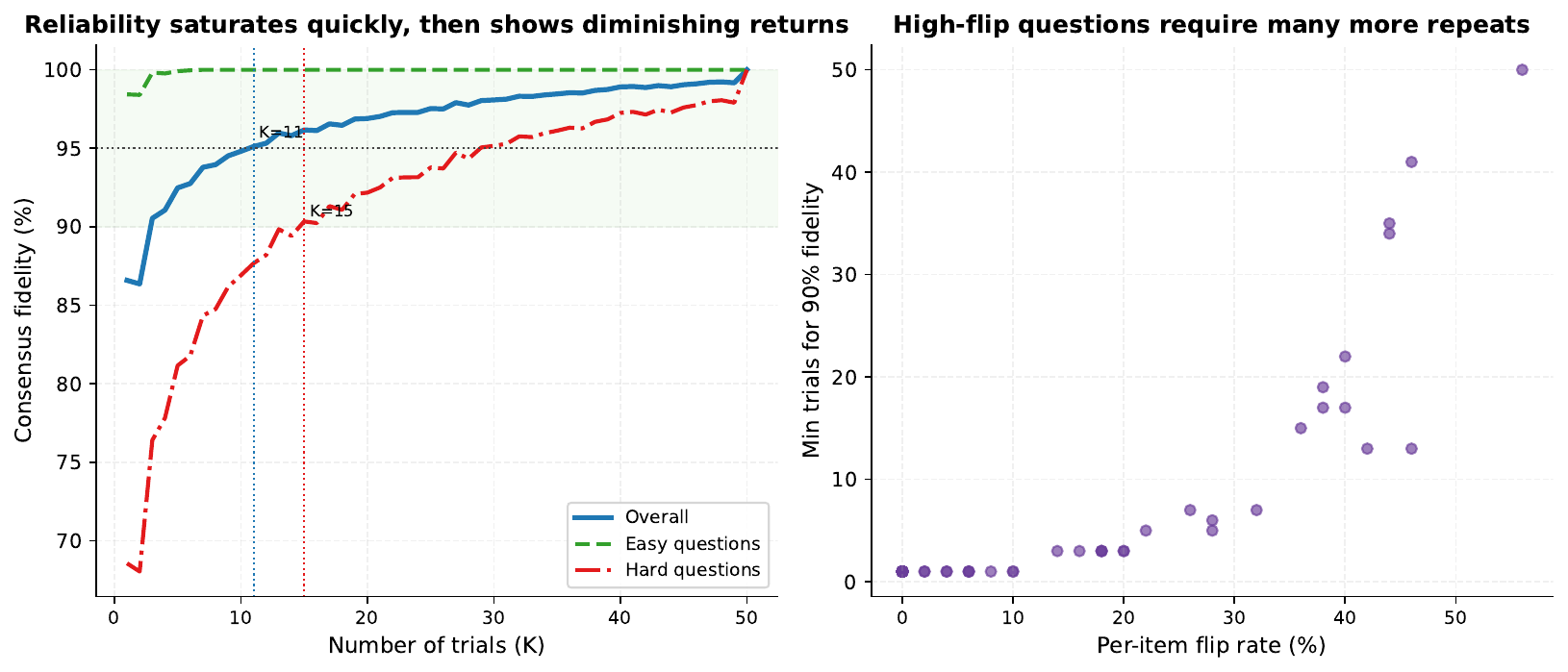}
    \caption{Reliability as a function of repeated voting. Left: probability that a $K$-trial majority vote matches the 50-trial reference verdict, shown overall and for easy vs.\ hard questions. Right: per-item minimum trial counts needed to reach 90\% fidelity, showing that high-flip items are disproportionately costly to stabilize.}
    \label{fig:reliability}
\end{figure}

The flip rate analysis quantifies \textit{how inconsistent} a judge is; a complementary question is \textit{how many trials are needed} to reach a stable verdict. Figure~\ref{fig:reliability} shows the reliability curve: $P(\text{majority}(K) = \text{majority}(50))$ as a function of $K$. As predicted by H3, the curve is sharply concave: repeated voting improves reliability quickly in the first few trials, then exhibits diminishing returns.

\begin{table}[h]
\centering
\caption{Minimum number of trials $K$ to reach 90\% and 95\% consensus reliability, overall and by difficulty stratum.}
\label{tab:minn}
\begin{tabular}{lccc}
\toprule
\textbf{Stratum} & \textbf{Min-$K$ (90\%)} & \textbf{Min-$K$ (95\%)} & \textbf{Note} \\
\midrule
All questions    & 3  & 11 & averaged across both judges \\
Easy (FR $<$ 10\%) & 1  & 3  & near-deterministic questions \\
Hard (FR $\geq$ 10\%) & 15 & $>$50 & high-variance questions \\
\bottomrule
\end{tabular}
\end{table}

A single trial achieves only 86.6\% consensus fidelity. Reaching 90\% requires approximately 3 trials on average, and 95\% requires 11 trials. Critically, these averages mask substantial stratification: for the 15 high-flip-rate questions (FR $\geq 10\%$), 15 trials are needed for 90\% fidelity, and 50 trials are still insufficient for 95\% on the hardest questions. The practical implication is direct: \textbf{single-trial LLM judge evaluations should be treated as preliminary estimates, not definitive verdicts}, particularly for questions in subjective or open-ended categories. For evaluation pipelines requiring high confidence, we recommend a minimum of 10--20 trials with majority voting; for adversarial or high-stakes comparisons (e.g., model release decisions), 50 trials may be warranted for borderline questions.

\subsection{Intraclass Correlation and Score Variance Decomposition}
\label{sec:icc}

\begin{table}[h]
\centering
\caption{ICC(2,1) (absolute agreement, single measures) for pointwise scores across 50 trials. ICC $< 0.60$ is conventionally ``poor to moderate'' reliability; ICC $0.60$--$0.75$ is ``moderate to good.''}
\label{tab:icc}
\begin{tabular}{lccc}
\toprule
\textbf{Judge} & \textbf{ICC (Response A)} & \textbf{ICC (Response B)} & \textbf{ICC (Combined)} \\
\midrule
GPT-4o-mini  & 0.608 & 0.533 & 0.575 \\
GPT-4.1-mini & 0.772 & 0.781 & 0.774 \\
\bottomrule
\end{tabular}
\end{table}

Table~\ref{tab:icc} reports ICC(2,1) for the 50-trial pointwise score sequences. GPT-4.1-mini achieves moderate-to-good reliability (ICC $= 0.77$), while GPT-4o-mini falls in the poor-to-moderate range (ICC $= 0.58$). Both values are substantially below the $\kappa > 0.80$ threshold typically required for high-stakes annotation tasks in clinical or legal settings.

To further characterize the nature of score variance, we decompose total pointwise score variance into between-question and within-question components. Across all 29 questions and both judges, 55.3\% of variance is between-question (reflecting genuine quality differences between the evaluated responses) while \textbf{44.7\% is within-question noise}, that is, variance attributable to the judge's stochastic response generation rather than to the quality of the responses being evaluated. This near-equal split is striking: for every point of meaningful signal in a pointwise score, there is almost an equal point of random noise. The ICC values are consistent with this decomposition: ICC $= 0.58$--$0.77$ means that 23--42\% of observed score variance is measurement error.

This variance decomposition has a direct implication for pointwise score interpretation: a single pointwise score of ``8 vs.\ 9'' cannot be reliably interpreted as evidence of a quality difference. With a within-question standard deviation of approximately $\sigma_w = \sqrt{0.359} \approx 0.60$ points, the 95\% margin of error for a single pointwise observation is $\pm 1.2$ points, large relative to the 0.19--0.36 mean score gaps observed between competitive responses.

\section{Discussion}

The results are easiest to interpret if we distinguish four layers of evaluation uncertainty. First, \emph{stochastic instability}: the same judge can change its verdict across repeated trials. Second, \emph{systematic bias}: for example, first-position preference can skew majority outcomes even when repeated trials are averaged. Third, \emph{protocol dependence}: temperature and prompt template alter the effective evaluator. Fourth, \emph{judge-identity dependence}: different judge models disagree on a non-trivial fraction of items. This layered view explains why single-trial LLM-as-a-Judge evaluations can appear deceptively crisp despite multiple underlying sources of variance.

\subsection{Comparison to Human Annotators}

A natural question is whether LLM judges are more or less consistent than human evaluators. Table~\ref{tab:human} contextualizes our findings against reported human baselines.

\begin{table}[h]
\centering
\caption{LLM judge consistency vs.\ reported human baselines.}
\label{tab:human}
\begin{tabular}{lcc}
\toprule
\textbf{Evaluator} & \textbf{Agreement} & \textbf{Source} \\
\midrule
Human (MT-Bench pairwise) & 81\% & \citet{zheng2023judging} \\
Human (Chatbot Arena) & 66\% & \citet{chiang2024chatbot} \\
Human (NLG subjective, $\kappa$) & 0.3--0.6 & \citet{amidei2019agreement} \\
\midrule
LLM intra-judge ($t=1.0$) & 86\%$^*$ & This work \\
LLM intra-judge ($t=0$) & 95\%$^*$ & This work \\
LLM inter-judge & 76\% ($\kappa=0.51$) & This work \\
\bottomrule
\multicolumn{3}{l}{\small $^*$Computed as $1 - \text{mean FR}$, averaged across both judges.}
\end{tabular}
\end{table}

At $t=1.0$, LLM intra-judge consistency (86\%) is comparable to reported human agreement ranges in some prior settings. However, inter-judge agreement (76\%) is lower than stronger human baselines from controlled settings. This nuance, namely that LLM judges can be individually consistent yet mutually inconsistent, is important: systematic error from judge choice may be more damaging to benchmark validity than stochastic error within a single judge.

\subsection{The Leaderboard Noise Budget}

A useful way to translate reliability statistics into benchmark design language is to ask how much avoidable label noise a judging protocol injects into a leaderboard. We call this quantity the \emph{noise budget}: the expected number of question-level outcomes that would change under repeated evaluation.

Our findings have a direct quantitative implication for benchmark validity. Define the \textit{noise budget} of a benchmark as the expected number of question-level evaluation outcomes that would change if the benchmark were re-run with a different random seed (or different judge model). Under single-trial judging at $t=1.0$ with mean flip rate 13.6\%, a 100-question benchmark has an expected noise budget of \textbf{13.6 incorrect outcomes per run}. For benchmarks with a score gap of $\leq$10 points between adjacent-ranked models, this is sufficient to reverse rankings with non-trivial probability.

This framing connects our findings to the practical benchmark design question. The noise budget shrinks substantially with multi-trial aggregation:

\begin{itemize}[nosep]
    \item \textbf{1 trial}: $\sim$13.6\% of outcomes incorrect (flip rate = mean FR)
    \item \textbf{3 trials}: $\sim$10\% of outcomes incorrect (P(correct) = 0.90 overall)
    \item \textbf{11 trials}: $\sim$5\% of outcomes incorrect (P(correct) = 0.95 overall)
    \item \textbf{20 trials}: $\sim$3\% of outcomes incorrect (P(correct) = 0.97)
\end{itemize}

For high-flip-rate questions (28\% of our dataset), the noise budget is much larger and requires 15+ trials for 90\% fidelity. Benchmark designers should therefore adopt differentiated trial counts: a quick initial screen with 5--10 trials to identify borderline comparisons, followed by targeted 20--50 trial evaluation for ambiguous cases.

\subsection{Implications for Evaluation Practice}
\label{sec:discussion_recs}

The main practical lesson is not merely that LLM judges are noisy, but that different forms of noise call for different countermeasures. Multi-trial voting addresses stochastic instability; position randomization addresses systematic order effects; multi-judge panels address judge-identity dependence; and prompt audits address protocol dependence.

Our findings have several concrete implications:

\textbf{Single-trial evaluations should be deprecated for publication-quality comparisons.} With a 14\% mean flip rate, one in seven pairwise comparisons changes outcome on re-run. The 86.6\% single-trial consensus fidelity is lower than the 81\% MT-Bench human agreement benchmark, which itself is considered marginal. Leaderboards and benchmarks should mandate multi-trial evaluation and report confidence intervals.

\textbf{Position must be randomized \textit{and} the randomization reported.} The 72\% A-wins position bias for GPT-4o-mini ($p = 0.024$) means fixed-position evaluation introduces systematic error. Even with randomization, the \textit{number} of position-randomized trials should be reported to allow variance estimation.

\textbf{Pairwise preferences should always accompany pointwise scores.} The pairwise-pointwise paradox (Section~\ref{sec:results_paradox}) is a fundamental problem with forced-choice formats: they generate spurious certainty when the underlying quality difference is below the judge's discrimination threshold. Dual reporting allows readers to assess whether pairwise differences are grounded in genuine quality gaps.

\textbf{Score variance is not negligible and should be reported.} ICC $= 0.58$--$0.77$ and 44.7\% within-question noise mean that a single pointwise score carries a 95\% margin of error of $\pm 1.2$ points on a 10-point scale. Reporting scores without confidence intervals misrepresents the precision of LLM judge evaluations.

\textbf{Judge selection is a confound, not a free choice.} The $\kappa = 0.51$ cross-judge agreement means that one in four evaluation outcomes depends on which judge is selected. Papers should report results across multiple judge models, or explicitly acknowledge judge selection as a potential confound.

\textbf{Prompt wording is a hidden experimental variable.} Semantically equivalent prompts change majority outcomes 25\% of the time. Evaluation papers should either standardize to community-accepted prompt templates or conduct prompt sensitivity analyses.

\textbf{Deterministic decoding ($t=0$) is necessary but not sufficient.} While $t=0$ reduces flip rates by 43--79\%, residual non-determinism at the API level means that even $t=0$ evaluations benefit from 3--5 trial repetitions.

\subsection{Recommendations}
\label{sec:recommendations}

Based on our findings, we propose a tiered evaluation protocol:

\begin{enumerate}[nosep]
    \item \textbf{Minimum standard (reproducibility)}: $\geq$10 trials at $t=0$ with randomized response order; report majority vote, flip rate per question, and question-level confidence intervals. This recommendation is grounded in the reliability curve: single-trial judging achieves only 86.6\% consensus fidelity, whereas 11 trials reach 95\% on average.
    \item \textbf{Standard practice (publication)}: 20 trials at $t=0$; dual-mode evaluation (pairwise + pointwise); multi-judge panel ($\geq$2 judges); report Cohen's $\kappa$ and ICC. This is motivated by the pairwise--pointwise gap and the cross-judge agreement result ($\kappa = 0.51$), which show that neither pairwise winners nor single-judge evaluations are sufficient on their own.
    \item \textbf{High-stakes evaluation (leaderboard / model release)}: 50 trials; identify high-flip-rate questions (FR $>$ 20\%) and flag them as ``uncertain''; use at least two judges from different providers; report noise budget alongside final scores. This follows from the hard-question regime, where FR rises to 23.6\% on average and 15 trials are needed just to reach 90\% fidelity.
    \item \textbf{Category-stratified reporting}: Report consistency metrics by task category, as reliability varies substantially across task types ($\eta^2 \approx 0.31$ for category on item-level flip rates), with coding and reasoning much less stable than knowledge and roleplay.
\end{enumerate}

\subsection{Limitations}

Several limitations should shape how broadly these results are interpreted. Most importantly, this is a careful repeated-trial study of a narrow slice of the LLM judge design space, not a universal audit of all judge models or all evaluation protocols.

Our study has several limitations. First, and most importantly, we evaluate only two judge models from the same provider (OpenAI). While they represent different model generations (GPT-4o vs.\ GPT-4.1 family), all findings are potentially artifacts of OpenAI's specific RLHF and decoding pipeline. Extending to other providers (Anthropic Claude, Google Gemini, open-source Llama/Mistral models) is necessary to establish generalizability, and we consider this the primary direction for future work. Second, our 29-question dataset, while diverse across 10 categories, is relatively small; the lack of statistical significance in category-level comparisons reflects this limitation. Third, while we test two prompt templates, the space of possible prompt designs is vast; further systematic exploration may reveal additional sensitivity patterns. Fourth, our competitive response pairs (both high-quality) represent a stress test, so consistency may be higher for response pairs with more obvious quality differences. Fifth, our response pairs use GPT-4o-mini (Response A) and GPT-4o (Response B), meaning GPT-4o-mini judges evaluate responses from their own model family. This introduces a potential self-preference confound that could partially explain the 72\% A-wins position bias observed for GPT-4o-mini; disentangling self-preference from genuine position effects requires a future controlled study using responses from out-of-family models. Finally, we do not directly compare to human annotators on the same task instances; our human baseline comparisons rely on reported values from prior work.

\section{Conclusion}

We present a repeated-trial study of LLM-as-a-Judge reliability across over 10,000 judgments. The core conclusion is not that LLM judges are useless, nor that they behave like random coin flips overall. Rather, it is that their reliability is layered and uneven: many items are judged stably, but a substantial minority remain noisy enough that single-trial evaluation is hard to justify.

Our results support three main takeaways. First, pairwise judgments are often unstable in precisely the kinds of subjective or competitive cases that matter most for benchmarking. Second, pairwise verdicts can overstate evidence for quality differences when pointwise scores remain nearly indistinguishable. Third, repeated majority voting improves reliability quickly but with diminishing returns, making multi-trial aggregation a practical and principled remedy.

We present the most comprehensive intra-judge consistency study of LLM-as-a-Judge evaluation to date, contributing six interconnected findings from over 10,000 judgments at 50 trials per question:

\begin{enumerate}[nosep]
    \item \textbf{14\% mean flip rate} (max 56\%), with 28\% of questions exceeding 20\%, enough noise to reverse close benchmark rankings.
    \item \textbf{Significant position bias} ($p = 0.024$, sign test), with 72\% of questions showing first-position preference for GPT-4o-mini.
    \item \textbf{Pairwise-pointwise paradox}: judges declare confident winners when pointwise scores are statistically indistinguishable ($p > 0.1$), revealing that pairwise forced-choice amplifies non-existent quality differences.
    \item \textbf{ICC(2,1) = 0.58--0.77} with 44.7\% of score variance attributable to within-judge noise, establishing that LLM judge pointwise scores have ``poor to moderate'' reliability by psychometric standards.
    \item \textbf{Reliability curve}: 11 trials are needed to reach 95\% consensus fidelity overall; 15+ for high-variance questions. Single-trial evaluations achieve only 86.6\% fidelity.
    \item \textbf{Multi-source variance}: temperature ($t=0$ reduces but does not eliminate inconsistency), prompt wording (25\% outcome flips), and judge selection ($\kappa = 0.51$) independently contribute to evaluation noise.
\end{enumerate}

These findings collectively establish a \textit{noise budget} for LLM-judge evaluation: a 100-question single-trial benchmark is expected to contain $\sim$14 incorrect pairwise outcomes, which shrinks to $\sim$5 with 11-trial majority voting. We propose a tiered evaluation protocol (Section~\ref{sec:discussion_recs}) calibrated to these reliability levels.

An important limitation is that both judges are from a single provider (OpenAI); replicating this analysis with Anthropic Claude, Google Gemini, and open-source models (Llama, Mistral) is the primary future work direction, as findings may not generalize across providers. Scaling to 100+ questions per category would also enable statistically powered category-level comparisons.

As the AI community increasingly relies on automated evaluation for model development, RLHF training, and public leaderboards, the reliability of the evaluator is not a secondary concern; it is a first-order validity question. We hope this work provides both the empirical grounding and the practical tools to elevate evaluation standards.

All data, code, and results are publicly available at \url{https://github.com/Abelo9996/llm-judge-consistency}.

\appendix

\section{Experiment Summary}

Table~\ref{tab:exp_summary} summarizes the experimental components used in this paper.

\begin{table}[h]
\centering
\caption{Summary of experimental components.}
\label{tab:exp_summary}
\begin{tabular}{p{0.20\textwidth}p{0.17\textwidth}p{0.23\textwidth}p{0.22\textwidth}}
\toprule
\textbf{Experiment} & \textbf{Questions} & \textbf{Trials} & \textbf{Purpose} \\
\midrule
Main pairwise + pointwise study & 29 questions, 2 judges & 50 pairwise trials + 50 pointwise trials per response & Estimate flip rate, pointwise gaps, ICC, and cross-judge agreement \\
Temperature ablation & 29 questions, 2 judges & 10 pairwise trials at $t=0$ & Measure how much deterministic decoding reduces instability \\
Prompt sensitivity study & 10 questions, 2 judges, 2 prompts & 20 pairwise trials per prompt & Measure protocol dependence under semantically equivalent prompt wording \\
Reliability simulation & Derived from main experiment & Monte Carlo subsampling over $K=1,\dots,50$ & Estimate how many repeated trials are needed for stable majority voting \\
\bottomrule
\end{tabular}
\end{table}

\section{Prompt Templates}

The prompt-sensitivity experiment compares two semantically equivalent templates.

\subsection{Prompt A}
\begin{quote}\small
\texttt{You are an impartial judge evaluating two responses to the same user request. Compare Response A and Response B on overall quality, helpfulness, correctness, and completeness. Return exactly one label: A, B, or tie.}
\end{quote}

\subsection{Prompt B}
\begin{quote}\small
\texttt{Please act as a fair and unbiased evaluator. Review both candidate responses carefully, think step by step about which one better satisfies the user request, and then output exactly one verdict: A, B, or tie.}
\end{quote}

Both prompts ask for the same pairwise decision, but they differ in framing, tone, and explicitness of instruction. The prompt-sensitivity analysis in Section~\ref{fig:prompt} measures how much those wording changes affect instability and majority outcomes.

\section{Per-Question Flip Rates and Majorities}

Tables~\ref{tab:perq_4omini} and~\ref{tab:perq_41mini} list per-question flip rates and majority outcomes for the two judges.

\begin{longtable}{llll}
\caption{Per-question pairwise outcomes for GPT-4o-mini.}\label{tab:perq_4omini}\\
\toprule
Question & Category & Flip rate & Majority \\
\midrule
\endfirsthead
\toprule
Question & Category & Flip rate & Majority \\
\midrule
\endhead
q001 & writing & 22\% & B \\
q002 & writing & 18\% & A \\
q003 & writing & 44\% & B \\
q004 & reasoning & 26\% & B \\
q005 & reasoning & 0\% & B \\
q006 & reasoning & 6\% & B \\
q007 & coding & 46\% & A \\
q008 & coding & 44\% & A \\
q009 & coding & 28\% & A \\
q010 & knowledge & 0\% & A \\
q011 & knowledge & 0\% & A \\
q012 & knowledge & 4\% & A \\
q013 & math & 0\% & A \\
q014 & math & 46\% & B \\
q015 & math & 14\% & A \\
q016 & roleplay & 0\% & A \\
q017 & roleplay & 0\% & A \\
q018 & extraction & 6\% & A \\
q019 & extraction & 2\% & B \\
q020 & extraction & 4\% & A \\
q021 & ethics & 0\% & A \\
q022 & ethics & 0\% & A \\
q023 & instruction & 18\% & A \\
q024 & instruction & 36\% & A \\
q025 & instruction & 0\% & A \\
q026 & hard & 0\% & A \\
q027 & hard & 20\% & B \\
q028 & hard & 0\% & A \\
q029 & hard & 2\% & A \\
\bottomrule
\end{longtable}

\begin{longtable}{llll}
\caption{Per-question pairwise outcomes for GPT-4.1-mini.}\label{tab:perq_41mini}\\
\toprule
Question & Category & Flip rate & Majority \\
\midrule
\endfirsthead
\toprule
Question & Category & Flip rate & Majority \\
\midrule
\endhead
q001 & writing & 0\% & B \\
q002 & writing & 40\% & B \\
q003 & writing & 0\% & A \\
q004 & reasoning & 56\% & tie \\
q005 & reasoning & 0\% & B \\
q006 & reasoning & 40\% & B \\
q007 & coding & 20\% & tie \\
q008 & coding & 38\% & B \\
q009 & coding & 8\% & tie \\
q010 & knowledge & 0\% & A \\
q011 & knowledge & 0\% & A \\
q012 & knowledge & 0\% & A \\
q013 & math & 18\% & A \\
q014 & math & 0\% & B \\
q015 & math & 42\% & A \\
q016 & roleplay & 0\% & A \\
q017 & roleplay & 0\% & A \\
q018 & extraction & 0\% & A \\
q019 & extraction & 32\% & B \\
q020 & extraction & 28\% & A \\
q021 & ethics & 16\% & A \\
q022 & ethics & 38\% & A \\
q023 & instruction & 0\% & A \\
q024 & instruction & 0\% & A \\
q025 & instruction & 6\% & A \\
q026 & hard & 0\% & A \\
q027 & hard & 10\% & B \\
q028 & hard & 10\% & B \\
q029 & hard & 0\% & A \\
\bottomrule
\end{longtable}

\section{ICC and Variance Decomposition Details}

Our pointwise-score reliability analysis uses ICC(2,1), the two-way random-effects, absolute-agreement, single-measures form of the intraclass correlation coefficient. This choice treats repeated judge calls as interchangeable raters and asks how much of the observed variance is attributable to stable between-item differences rather than within-item stochasticity.

In the combined variance decomposition, 55.3\% of pointwise-score variance is between-question signal and 44.7\% is within-question noise. This near-even split is the main reason single pointwise scores should be interpreted cautiously: the stochastic component is too large to be ignored when score differences are small.

\bibliography{references}

\end{document}